# Phonetic Segmentation of the UCLA Phonetics Lab Archive

**Eleanor Chodroff[1], Blaž Pažon[2], Annie Baker[1], and Steven Moran[3,4]**

[1]University of Zurich, [2]University of Ljubljana, [3]University of Neuchâtel, [4]University of Miami
[1]Department of Computational Linguistics, Andreasstrasse 15, 8050 Zürich, Switzerland
{eleanor.chodroff, annie.baker}@uzh.ch, blaz.pazon@gmail.com, steven.moran@unine.ch

## Abstract

Research in speech technologies and comparative linguistics depends on access to diverse and accessible speech data. The UCLA Phonetics Lab Archive is one of the earliest multilingual speech corpora, with long-form audio recordings and phonetic transcriptions for 314 languages (Ladefoged et al., 2009). Recently, 95 of these languages were time-aligned with word-level phonetic transcriptions (Li et al., 2021). Here we present VoxAngeles, a corpus of audited phonetic transcriptions and phone-level alignments of the UCLA Phonetics Lab Archive, which uses the 95-language CMU re-release as our starting point. VoxAngeles also includes word- and phone-level segmentations from the original UCLA corpus, as well as phonetic measurements of word and phone durations, vowel formants, and vowel f0. This corpus enhances the usability of the original data, particularly for quantitative phonetic typology, as demonstrated through a case study of vowel intrinsic f0. We also discuss the utility of the VoxAngeles corpus for general research and pedagogy in crosslinguistic phonetics, as well as for low-resource and multilingual speech technologies. VoxAngeles is free to download and use under a CC-BY-NC 4.0 license.

## 1. Introduction

Advancements in speech technologies and comparative linguistics research benefit from increased availability of accessible crosslinguistic speech data. Goals ranging from language description and preservation, phonetic theory and typology, to automatic speech recognition or text-to-speech synthesis, require diverse crosslinguistic data. While "big data" has allowed incredible development in both speech science and technology for high-resource languages, the quantity of such data simply does not exist for the majority of the world's languages. Moreover, for speech data that might exist, it is unlikely to be in a directly usable format for many research questions in industry or science.

The usability of a speech corpus depends considerably on the research question. The mere existence of speech audio data may suffice for some research questions, whereas for others, metadata may be necessary for downstream analysis. The most useful accompanying metadata is a written orthographic and/or phonetic transcription of the spoken audio which can enable supervised training of speech technology systems, as well as the searchability of the audio for key phrases or segments.

Access to crosslinguistic speech corpora has risen dramatically in recent years, particularly with the release of several massively multilingual speech corpora. This growth spurt has arisen in large part from technological advances in computational power including increased storage and processing capacity, as well as the increased accessibility of decent recording devices through personal smartphones or computers. Two such massively multilingual speech corpora are the CMU Wilderness Corpus with approximately 20 hours of speech per language for over 600 languages (Black, 2019), and the Mozilla Common Voice Dataset with anywhere from 1 to over 3000 hours of speech per language for over 100 languages (Ardila et al., 2020). The direct usability of these corpora for many researchers comes through the availability of utterance-level alignments, as well as the development of automatic word- and phone-level alignments for many of the languages (VoxClamantis for Wilderness: Salesky et al., 2020; VoxCommunis for Common Voice: Ahn and Chodroff, 2022; CommonVoice Utils: Tyers and Meyer, 2021).

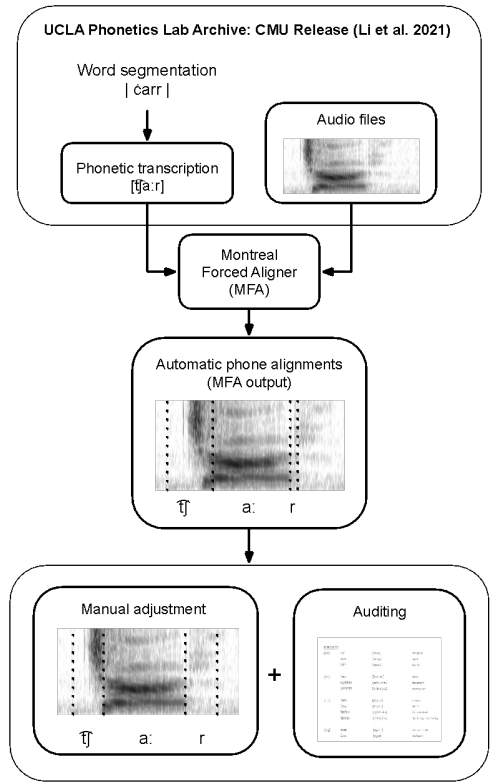

Figure 1: The workflow for aligning and auditing the CMU release of the UCLA Phonetics Lab Archive for inclusion in the VoxAngeles corpus.

Additional but smaller-scale crosslinguistic speech corpora have also been beneficial to the speech community, particularly due to the availability of utterance-, word-, or phone-level alignments for downstream acoustic model training and phonetic analysis. These include, but are not limited to the GlobalPhone Corpus with 20 languages (Schultz et al., 2013), Multilingual LibriSpeech with eight languages (Pratap et al., 2020), and the DoReCo Corpus of fieldwork recordings from over 50 languages (Seifart et al., 2022).

Indeed, developing usable crosslinguistic speech corpora has long been a goal in the speech community. The UCLA Phonetics Lab Archive is one of the earliest, and to this day, one of the most crosslinguistically diverse collections of speech data (Ladefoged et al., 2009). The corpus contains spoken audio recordings and phonetic transcriptions from 314 languages representing 51 language families, and is freely available to the public in an online repository.[1] The data were collected in fieldwork sessions from the 1960s to the late 1990s (Ladefoged and Maddieson, 1996a). Between 2006 and 2009, a web page was created for each language, which contains links to the original audio recordings and fieldnotes, along with digitized tables of the phonetic transcriptions, orthographic forms, and in many cases, translations (Ladefoged and Schuh, 2004; Ladefoged et al. 2009). Though the corpus includes phonetic transcriptions of the target language, the transcriptions have not previously been time-aligned to the recordings.

In a large and valuable undertaking, Li et al. (2021) obtained time-aligned transcripts for recordings from 95 languages of the original corpus using an automated approach. Specifically, the word-level phonetic transcriptions were scraped from the primary online database, and the long audio files were segmented into individual recordings of transcribed words. Through this process, utterances in the non-target language (e.g., the language of interaction during the fieldwork session) could be bypassed. The audio segmentations were then validated by a human listener to ensure the alignment between the text and audio was correct. This effort greatly enhanced the usability of the corpus, enabling direct work on multilingual and universal phone recognition (e.g., Li et al., 2022; Liu et al., 2023).

In the present paper, we present VoxAngeles[2], an updated release of the UCLA Phonetics Lab Archive, that includes manually corrected phone-level alignments using the original, or closely adapted phonetic transcriptions, as well as phonetic measurements of phone and word durations, vowel formants, and vowel f0. In total, the current release of the corpus spans 95 languages from 21 language families (as defined by Glottolog; Hammarström et al., 2023). In the final portion of the paper, we demonstrate the utility of this corpus for phonetic typology in a case study of a previously posited crosslinguistic phonetic universal, vowel intrinsic f0.

## 2.  Methods

The overall procedure for developing the VoxAngeles corpus is depicted in Figure 1. Using the word segmentations and extracted phonetic transcriptions from the 95 languages in the CMU release as input (Li et al., 2021), we automatically force aligned the phonetic transcription to the audio file using the Montreal Forced Aligner (McAuliffe et al., 2017; Section 2.1). The phone alignments were then manually adjusted and audited along several dimensions (Section 2.2), and additional data were segmented from the original UCLA Phonetics Lab Archive where possible (Section 2.3). Finally, acoustic-phonetic measurements were extracted for phonetic analysis (Section 2.4).

### 2.1  Phonetic Forced Alignment

The CMU release of the UCLA Phonetics Archive contains individual audio files from 95 languages, in which each audio file is accompanied with a word-level phonetic transcription scraped from the original corpus website (Li et al. 2021). The phonetic transcriptions were largely based on the IPA, though several discrepancies are outlined in Section 2.2. To obtain phone-level time alignments, the files were prepared for phonetic forced alignment using the Montreal Forced Aligner (MFA). We employed the most diverse and at the time, largest, pretrained acoustic model available through the MFA. This was the "english 2.0.0a" acoustic model, which we refer to as the "Global English model" (McAuliffe and Sonderegger, 2022). It was trained on 3700+ hours of global English, including recordings from American, British, Nigerian, South African, and Irish varieties of English, among others. This model has been shown to be consistently competitive with or more accurate than small language-specific acoustic models (~1 hour) in forced alignment performance (Chodroff et al., 2024).

As one would expect, the phonetic symbols used in the UCLA Phonetics Lab Archive show remarkable diversity (see also Maddieson, 1984; Moran and Cysouw, 2018). For alignment with the pretrained Global English model, we needed to remap these phones to those in the acoustic model phone set. Thus, we identified phonetically comparable segments for the remapping process, and used the Interlingual MFA toolkit (Dolatian, 2023) to maintain a mapping between the original and English-adapted phonetic transcriptions.

The initial alignment was then implemented using the MFA, the Global English acoustic model, and the custom pronunciation lexicon with the remapped UCLA Phonetics Lab Archive transcriptions. Following alignment, the phones were reconverted back to their original transcription using the Interlingual MFA toolkit (Dolatian, 2023). The resulting set of Praat TextGrids with the word- and phone-level alignments were then distributed to two phonetically

---

trained annotators for manual adjustment and auditing.

## 2.2 Manual Alignment and Auditing

During the process of manual alignment, we adhered to three main guiding principles. Firstly, we aimed to remain faithful to the transcription produced by the linguist who originally compiled and transcribed the word list. Second, we aimed to represent the entire speech signal with the transcription provided, and third, to assign a section of the speech signal to each element of the transcription.

Despite the overall intention to stay faithful to the original transcription, in some cases it was considered beneficial to modify those in the MFA output TextGrids. This was the case for transcriptions which featured obsolete IPA symbols, errors occurring at different points of the extraction process, or the use of non-standard phonetic symbols. These modifications were implemented in the interest of maintaining consistency throughout the dataset. Some of the more common issues or obstacles we encountered included:

**Inconsistent representation of suprasegmental features.** The primary exception to the fundamental guideline of retaining the original transcription was the marking of tones, syllable boundaries, and stress. Where included, these tended to be marked inconsistently throughout the corpus, which in some cases hindered automatic processing. These were therefore removed.

**Obsolete and nonstandard symbols.** As the UCLA Phonetics Lab Archive compiles work by various investigators conducted over several decades, there was an unsurprising lack of uniformity when it came to the employed set of phonetic symbols. Some of these symbols were obsolete IPA symbols, some from the Americanist phonetic notation, and some perhaps were developed to simplify the transcription in handwriting or with a typewriter. We aimed to include only current standard IPA symbols, and to ensure consistency across languages. To standardize the notation, we frequently consulted the scanned copies of the original field notes. Many of the notes included legends that mapped a simplified symbol to a more complex phonetic transcription.

First, several of the IPA symbols used in the original transcriptions are now obsolete. Among the more common were <ɩ> (lowercase iota), representing a near-close near-front unrounded vowel, replaced by the current standard <ɪ> (small capital I), and <ɷ> (closed omega), representing a near-close near-back rounded vowel, replaced by the current standard <ʊ> (inverted omega). As of the 1989 Kiel Convention, however, the symbols [ɩ ɷ] were standardized to [ɪ ʊ], which remains the standard to this day (Roach, 1989). Many of the obsolete symbols were also missing in the extracted TextGrids, likely from rendering issues in the digitized HTML file.

Cases in which there was no clear one-to-one correspondence between the original symbol and a modern equivalent were addressed by referring to, where available, relevant literature on the language in question. For example, in the case of Angami (njm), certain segments in the transcriptions did not have a clear modern equivalent. In that case, we were able to consult a phonetic description of the grammar in Blankenship et al. (1992), which allowed us to update the MFA output TextGrid transcription from [whe] to [ʍe], and from [mhe] to [mʰe].

In some cases, nonstandard symbols, frequently but not always from the North American Phonetic Alphabet, were adapted to current standard IPA symbols to ensure consistency and maintain the necessary phonetic contrasts. An example of this is Ladino (lad), which distinguishes an alveolar tap /ɾ/ from an alveolar trill /r/; these were represented in the transcripts respectively by <r> and <ŕ>. The latter is not standard IPA, so we changed all original <r> symbols to <r>, and <ŕ> to <r>.

Additional common modifications are listed in the following. Many of these mappings were determined based on the identification of a legend in the original field notes: <ü> replaced by <y>, <ä> replaced by <æ>, <š> replaced by <ʃ>, <č> replaced by <tʃ>, <ž> replaced by <ʒ>, <ǯ> replaced by <dʒ>, <ñ> or <ñ> replaced by <ɲ>, <ʎ> replaced by <tɬ>, and the underdot denoting retroflex consonants (e.g., <ṭ>) replaced by the modern retroflex consonant symbols (e.g., <ʈ>).

**Typographical issues.** The symbol <g> used on the UCLA website was in most cases rendered as <g̠> (underlined <q>) at some point during the automatic extraction process undertaken by Li et al. (2021).

In several cases, the remapping between the original and the English-adapted phonetic transcriptions for the MFA was not entirely successful, even in cases where the phone was not remapped. The reason for this is unclear; however, in each case these were manually corrected during the auditing process. For example, while the symbol <ɲ> was preserved in the word tier, it was replaced by <ɲ> in the phones tier in Scottish Gaelic (gla), Gujarati (guj) and Kannada (kan). This occurred less frequently with other symbols and less consistently across different words. Other examples included: <g> replaced by <k> in Basque (eus) and Kannada (kan), <ʁʷ> replaced by <f> in Chamalal (cji), <d> replaced by <d> in Gujarati (guj), <ʌ> replaced by <ə> and <ɾ> replaced by <ɹ> in Kannada (kan).

**Missing symbols.** Some symbols exhibited rendering issues which led to complete disappearance, such as tie bars combining less frequent sound combinations (e.g., /hw/ in Bassa (bsq), or /p͡ʃ/ in Hebrew (heb)), various diacritics (including carons, underdots, modifier letters, symbols denoting nasalized, velarized, rounded and dental phones), and some less common IPA symbols (e.g., <ʢ> and <н> in Aghul (agx)). A status shared by all missing symbols was that of being in some way removed from the more frequently used and core IPA symbols; they were either non-standard, non-IPA or symbols which are rarely used, both independently or in certain uncommon combinations.

This can be observed for example in the case of the Lithuanian (lit) word *kelti* ('to lift'): both the field notes and the digitized word list on the UCLA Phonetics Lab Archive webpage provide the transcription [kʲæˑlʲtʲi], whereas the CMU word list and consequently the MFA output TextGrid have the transcription [kʲælʲtʲi], with missing diacritics and the obsolete <ɪ> symbol (discussed above).

**Unclear segment boundaries.** As with the majority of segmentation tasks, there were cases in which boundaries between two segments were difficult to define (see for example Figure 2). In our attempt to remain faithful to the original transcription, we made sure to insert a boundary, even when the acoustics did not strongly indicate a specific point of separation. The absence of clear boundaries is to be expected, considering that speech does not consist of isolated sounds; rather, it manifests as a continuous speech signal shaped by coarticulation.

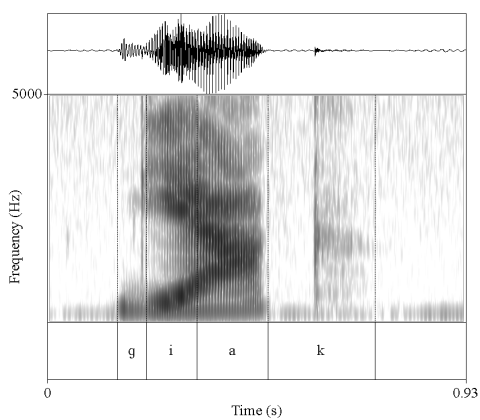

Figure 2: Example of unclear boundaries in a case of adjacent vowels with dynamic formant trajectories from Min Nan (nan).

**Transcription–audio mismatch.** For multiple files the phonetic transcription provided in the MFA output TextGrids did not exactly match the audio. This presented a problem when aligning, and was addressed by consulting the word list in the database, the original recording and field notes. This facilitated the attribution of the transcript–audio mismatch to one of a few possible errors. Usually, the error could be attributed to a mismatch of the transcript and the audio. If the corresponding transcript was found in the word list and we could confirm it to be the correct match for the audio, we simply adapted the transcript accordingly. In some cases, further inspection revealed the cause of the mismatch to be an inconsistency between the original transcript in the field notes and that of the digitized word list. In such cases the digitized word list typically contained an error and we opted for the original transcript. An example of this discrepancy is the transcript of the Danish (dan) word kantate 'cantata', transcribed as /kʰɛtʰiðə/ in the digitized word list, and /kʰɛntʰeiðə/ in the field notes. The latter was used, as it was not only the original transcript, but it also better represented the speech signal.

In other cases, the audio file itself was problematic. A prime example of this was a string of faulty audio files at the end of the dataset for Armenian (hye), where some audio files were mismatches, while the rest contained only noise from the original recording. Such recordings containing only noise also occurred in Ibibio (ibb), while one recording of Igbo (ibo) contained only the linguist's speech. Such mismatched files were removed.

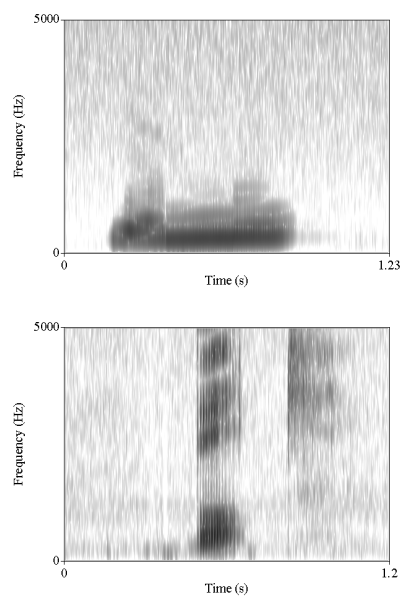

Figure 3: Example of audio quality differences between Malayalam (mal; top) and Dutch (nld; bottom). The spectrogram settings are the same for both images.

**Audio quality.** The overall quality of the audio recordings varied considerably, largely due to the wide range of years and environments over which the data were collected (see for example Figure 3). Users of the corpus should be aware of this disparity of quality between languages.

Occasionally, certain recordings were interrupted by loud noises, such as thuds or vocalizations from another speaker. Rarely, when a word-final consonant was pronounced silently or was a plosive with a long occlusion phase, the recording was cut off prematurely. In even rarer cases, the audio was cut off at the onset of the word. In each of these scenarios, the files were removed, as the information from the sounds was interrupted or incomplete.

## 2.3 Additional Data

Data from 11 additional languages were manually aligned at the word-level and extracted into individual recordings. These have also been force-aligned using the same procedure as described in Section 2.1, but have not yet gone through manual correction. In total, 1669 recordings were added to the corpus, with a median of 55 new recordings per language and a range of 23 to 692 new recordings: South Levantine Arabic (ajp): 92; Western Apache (apw): 40; South Azerbaijani (azb): 55; Bemba (bem): 23; Bengal (ben): 250; Albay Biscolano (bkh): 35; Edo (bin): 45; Bassa (bsq): 44; Czech (ces): 268; Chamorro (cha): 125; Degema (deg): 692.

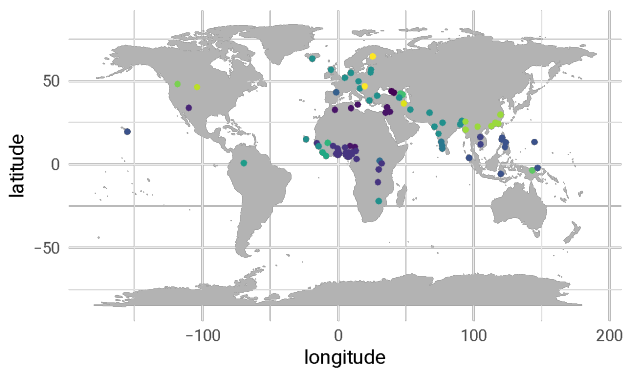

Figure 4: World map of the 95 aligned and audited languages from the UCLA Phonetics Lab Archive. Each color represents a distinct language family.

## 2.4   Acoustic-phonetic Analysis

The current release includes the following pre-extracted acoustic-phonetic measurements: the duration of all phone segments, the f0 at each quartile and decile of corner vowels, and F1–F3 at each quartile and decile of corner vowels. Along with each of these measurements, we include information about the preceding and following phones, the corresponding word, file name, the phone start and end times, and the word start and end times. The repository also includes the corresponding Praat scripts that can be modified to expand the targeted segment set.

All measurements were extracted using Praat. F0 contours were extracted using the To Pitch function, with the floor at 75 Hz, the ceiling at 500 Hz, and a time step of 0.01 s. Formant contours were extracted using the To Formant (burg) function, with parameters informed by the average vowel midpoint f0 of the file. If the average vowel midpoint f0 was over 160 Hz (the mean of our data), the formant ceiling was set to 5500 Hz. Otherwise, it was set to 5000 Hz. In all cases, the window length was 0.025 s, the time step was 0.01 s, the maximum number of formants estimated was 5, and pre-emphasis was added from 50 Hz.

## 3.   Results

The manual alignment yielded a total of 5,445 word-level recordings from 95 languages (Figure 4), with a median of 49 recordings and a range of 20 to 162 recordings per language. The languages span 21 language families (Hammarström et al., 2023; see Table 1). Within these files, a total of 22,825 phone intervals were aligned, with a median of 228 and a range of 46 to 755 phone intervals per language. Within the corpus, 568 distinct phones were present. The number of distinct phone types per language ranged from 13 to 93 with a median of 35. Of the observed phone types, approximately 209 were vowels, 184 were stops, affricates, trills, or taps, 94 were fricatives, 41 were approximants, and 40 were nasal consonants.

| Language family | # languages | # files |
|---|---|---|
| Atlantic-Congo | 27 | 922 |
| Indo-European | 21 | 1650 |
| Austronesian | 9 | 419 |
| Afro-Asiatic | 8 | 534 |
| Sino-Tibetan | 7 | 484 |
| Dravidian | 3 | 140 |
| Abkhaz-Adyge | 2 | 175 |
| Austroasiatic | 2 | 115 |
| Kru | 2 | 99 |
| Nakh-Daghestanian | 2 | 116 |
| Uralic | 2 | 229 |
| Athabaskan-Eyak-Tlingit | 1 | 62 |
| Basque | 1 | 46 |
| Central Sudanic | 1 | 23 |
| Ijoid | 1 | 85 |
| Kakua-Nukak | 1 | 63 |
| Mande | 1 | 69 |
| Nuclear Torricelli | 1 | 70 |
| Salishan | 1 | 46 |
| Siouan | 1 | 42 |
| Turkic | 1 | 56 |

Table 1: Language families in the audited subset of the corpus along with the total number of languages and word-level files per family.

Relative to the CMU release, a total of 64 files were removed from across 28 languages, mostly due to interfering noise (see Section 2.2). In close reference to the original field notes and for reasons outlined in Section 2.2, we also updated phonetic transcriptions for around 1,354 words from 85 languages. This corresponded to about 25% of the dataset. Within these languages, a median of 9 transcriptions were changed per language, with a mean of 16 and a maximum of 88.

Of the phone intervals retained in the comparison between the original MFA alignment and the corrected versions, the median onset boundary adjustment was 6.9 ms. Approximately 71% of the boundaries were within 20 ms of the gold boundary, 57% within 10 ms, and 45% within 5 ms. About 37% of the boundaries were unchanged. Overall, this indicates decent performance of the automatic forced alignment.

## 4. Case Study: Intrinsic f0

In its phonetically time-aligned format, the VoxAngeles corpus serves as a foundational resource for investigations in phonetic typology. Several phonetic universals have been previously suggested in the literature, but there has been limited empirical research due to the lack of access to diverse crosslinguistic speech data. Such universals include intrinsic f0 or intrinsic vowel duration, where low vowels (e.g., /a/ or /ae/) tend to have a lower f0 and longer vowel duration than corresponding high vowels (e.g., /i u/) (Meyer, 1896–7; Whalen and Levitt, 1995). These may be due to an automatic biomechanical consequence of the tongue or jaw movement, but with the underlying assumption that the intended phonetic target was uniform for each vowel category. The degree to which these effects are automatic or under speaker control has, however, been extensively debated in the field (Diehl and Kluender, 1989; Whalen and Levitt, 1995; Ting et al., 2023). While phonetically robust, these phenomena are rarely phonologized (cf., consonant f0 in which a high f0 after voiceless consonants and low f0 after voiced consonants can lead to tonogenesis; see Ting et al., 2023 for an overview).

Intrinsic f0 has had the benefit of being studied across a wide range of languages. Whalen and Levitt (1995) conducted a meta-analysis of the literature, identifying previously reported vowel-specific f0 means from approximately 31 languages across 11 language families. They found a significant difference between the f0 of high and low vowels across languages, with language-specific means largely demonstrating the expected numerical direction. More recently, Ting et al. (2023) investigated the presence and magnitude of this effect in a large-scale crosslinguistic study using 16 languages across 8 language families in the GlobalPhone and Librispeech Corpora (Schultz et al., 2013; Panayatov et al., 2015). They additionally confirmed the presence of the effect in each language, but to varying degrees of magnitude. In this study, we contribute to this literature by investigating whether high vowels such as /i/ and /u/ have a higher f0 than low vowels such as /a/, with increased language diversity in the dataset. Through this process, we can also assess the relative strengths and weaknesses of the VoxAngeles corpus for such investigations.

### 4.1 Methods

The midpoint f0 was extracted from high and low vowels using the extraction methods described in Section 2.4. First, vowel types with fewer than 10 tokens and devoiced vowels were excluded. Vowels were then collapsed into three broad categories: high front vowels [i iː i̝ ɪ ĩ ɨ iː y ʏ], high back vowels [u uː ʊ ŭ ʉ ɯ], and low vowels [æ æː a aː a̝ ɑ ɑː ɑ̃]. For simplicity, we will refer to these respectively as /i/, /u/, and /a/. Language-specific broad vowels with fewer than 10 tokens were excluded from analysis.

Midpoint f0 was converted to ERB (Glasberg and Moore, 1990), a perceptual scale, then averaged by broad category type within each language. As a descriptive analysis, we calculated the numerical difference of the effect on the averages, and whether the direction conformed to the previously posited universal at a descriptive level. Inferential analyses were conducted at two levels: the language-specific level and the cross-language level. For the language-specific level, we aimed to test the reliability of an intrinsic f0 effect within a language, and determine the proportion of languages that exhibited the expected relationship. The midpoint f0 in ERB was predicted from the broad vowel category of /i/ vs /a/ or /u/ vs /a/. Two models were chosen instead of one given that only one of the two high vowel categories passed our filtering criteria for many of the languages. Given the frequently small sample size, unbalanced contexts, and single speaker status, only simple linear regressions could be run on each language. Significance was assessed after applying the conservative Bonferroni correction, in which the alpha of 0.05 was divided by the total number of tested languages.

To investigate how reliably this effect held across languages, we submitted the raw data to two linear mixed-effects model with f0 in ERB as the dependent variable and fixed effects of broad vowel category (/i/ vs /a/ or /u/ vs /a/; treatment-coded), the voicing status of the preceding segment (voiced, voiceless, or silence; treatment-coded against voiced), the voicing status of the following segment (voiced, voiceless, or silence; treatment-coded against voiced), their interactions, the duration of the vowel, and random intercept and slopes for vowel category, preceding segment, and following segment by language. Only one speaker was observed per language so the language-specific random effects doubled as a speaker effect. The first model compared /i/ to /a/ in the broad vowel category, and the second compared /u/ to /a/. Effects with a t-value beyond an absolute value of 2.0 were taken as significant.

### 4.2 Results

After data exclusion, 53 languages from 17 language families were available for the /i/–/a/ analysis, and 36 languages from 13 families for the /u/–/a/ analysis.

Figure 5 illustrates the variability as well as the systematicity in midpoint f0 within a speaker and language for each comparison between the high and low vowels. Each point representing the paired f0 means lies quite close to the dashed line of identity. The correlation of paired mean f0s was significant for each high–low vowel comparison (/i/–/a/: $r = 0.95$; /u/–/a/: $r = 0.94$; each $p < 0.001$), demonstrating a highly predictable relationship between the f0 values of the two categories across languages.

For the /i/–/a/ comparison, 39 of the 53 languages conformed to the expectation with a higher f0 for /i/ than /a/ (74%), whereas 14 of the 53 languages showed the numerically opposite ranking. As can be seen in Figure 3, the differences between /i/ and /a/ f0s were generally small. After the Bonferroni correction, only four of the 39 direction-conforming languages reached significance and one of the 14 non-conforming languages. With an unadjusted alpha of 0.05, the number of "significant" differences reached 19 of the 39 conforming languages, and two of the 14 non-conforming languages. Though most languages indeed demonstrated the expected direction of the effect, the differences were generally small and not significant, at least with these sample sizes.

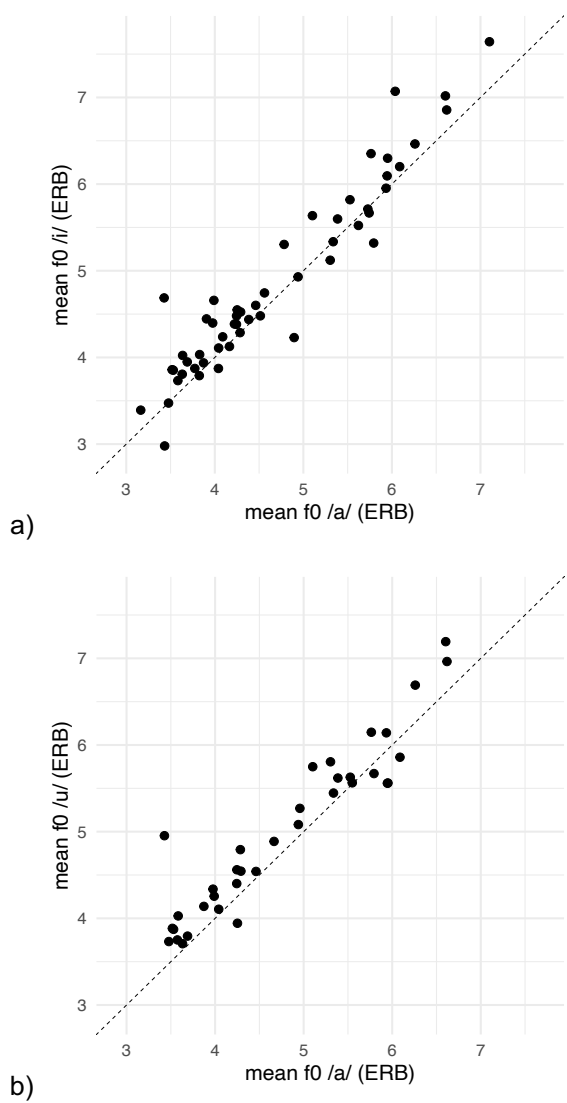

a)

b)

Figure 5: Paired language-specific mean midpoint f0s in ERB for: a) /i/ and /a/ and b) /u/ and /a/. The dashed line corresponds to the line of identity: points above the line reflect a higher f0 for the high vowel than the low vowel.

Similar patterns of findings were observed for the /u/–/a/ comparison: 32 of the 36 languages conformed to the expectation with a higher f0 of /u/ than /a/ (89%), whereas five languages showed the numerically opposite ranking. As before, the differences were generally small between /u/ and /a/, particularly if the f0 difference went in the opposite direction from expectations. With the Bonferroni-adjusted alpha value, a significant difference was observed for only four of the 31 direction-conforming languages and zero of the five non-conforming languages. With an unadjusted alpha of 0.05, 15 of the 31 conforming languages showed a significant difference, along with one of the five non-conforming languages.

The majority of the language-specific results were numerically, but not always significantly in line with the expected intrinsic f0 direction; however, the crosslinguistic analysis revealed a reliable effect of vowel height across languages for each of the /i/–/a/ and /u/–/a/ linear mixed-effects analyses (Table 2).

| Effect | /i/–/a/ | | /u/–/a/ | |
|---|---|---|---|---|
| | Est. β | *t*-val. | Est. β | *t*-val. |
| Vowel height | 0.24 | 4.33 | 0.16 | 2.15 |
| Preceding voiceless | 0.08 | 2.19 | 0.09 | 2.07 |
| Following silence | −0.20 | −4.88 | −0.17 | −3.69 |
| Vowel duration (100 ms change) | 0.08 | 4.54 | 0.05 | 2.60 |
| Height × preceding voiceless | *0.12* | *1.74* | 0.19 | 2.36 |
| Height × following voiceless | *0.04* | *0.52* | 0.20 | 2.14 |
| Height × following silence | −0.16 | −2.38 | *−0.03* | *−0.30* |

Table 2: Estimated beta values and *t*-values for significant effects in the /i/–/a/ or /u/–/a/ mixed-effects linear model of f0 (ERB). (Italicized numbers with reduced font size reflect non-significance in the indicated model.)

The patterns of significance for the main effects were consistent between the /i/–/a/ and /u/–/a/ models (see Table 2). A preceding voiceless segment, a following silence, and a longer vowel duration corresponded to a significantly higher f0 for both the high and low vowels. The presence of a preceding silence or following voiceless segment did not significantly influence overall vowel f0.

Regarding the interactions with vowel height, some minor differences emerged between the two models. In the /i/–/a/ model, the difference between /i/ and /a/ f0s significantly decreased following a voiceless segment and before silence. In the /u/–/a/ model, the difference in f0 between /u/ and /a/ following a voiceless segment, as well as before a voiceless segment was significantly larger. In both models, a preceding silence did not significantly influence the intrinsic f0 effect.

### 4.3 Discussion and Limitations

On a macro-level, the effect of vowel height on f0 was significant, and we confirmed previous findings that final boundaries can diminish the effect (e.g., the interaction of vowel height and following silence; Ladd and Silverman, 1984; Shadle, 1985). Moreover, we observed highly similar f0s across vowel heights, which is consistent with the idea that f0 should be near-uniform across high and low vowels, despite any automatic or even intentional f0 enhancement effects (automatic effects: Whalen et al., 1998; intentional effects: Diehl and Kluender, 1989; uniformity: Chodroff and Wilson, 2022). Nevertheless, the stability of the effect varied across languages and speakers. A few counterexamples in the direction of the effect were observed for certain speakers and languages: this could reflect inherent speaker- or language-specific variability of the phenomenon, or simply be an artifact of unbalanced phonetic environments or low sample sizes.

A few contradictory languages were also found relative to previous findings. In our sample, Finnish, Greek, and Hausa had numerically non-conforming f0 effects for /i/ and /a/ (/a/ f0 > /i/ f0). Whalen and Levitt (1995) found numerically expected differences for each of these languages, even with a low sample size (1–2 speakers). Ting et al. (2023) additionally found a significant f0 contrast in the expected direction across 100 speakers of Hausa, though speaker-specific deviations were not reported.

Overall, we observed crosslinguistic support for an intrinsic f0 effect, in which the f0 of high vowels tends to be higher than that of low vowels. Nevertheless, the findings were more variable across individual speakers and languages than has been previously reported. It may be that with a larger sample size per language and more favorable prosodic contexts, the effect could stabilize. Alternatively, previous studies could have been limited by a relatively smaller number of languages and language families; the effect could vary depending on the speaker or language. The findings call for further investigation of the effect at scale, across a more diverse set of languages, and with consideration of the effect's stability from speaker to speaker.

The analysis is indeed limited by shortcomings of the corpus: at the present, only data from one speaker per language is available, and only one word per utterance. These present unfortunate confounds between speaker and language, as well as between the measurement and prosodic position. Our future work aims to extract additional data from the UCLA Phonetics Lab Archive, which will marginally increase the number of speakers per language, as well as the overall sample size. We also hope to extract longer passages of spoken data when available. Finally, further extraction and segmentation from unprocessed languages will further diversify the available data for analysis.

### 5. Conclusion

The current paper presents VoxAngeles, an audited and phonetically aligned corpus of the UCLA Phonetics Lab Archive, supplemented with phonetic measurements and processing scripts. The development of this resource would not have been possible without the foundational work through NSF-funded fieldwork and endangered language documentation (Ladefoged and Maddieson, 1991, 1994, 2006a, 2006b; Ladefoged and Schuh, 2004; Ladefoged et al., 2009), as well as the automatic word-level alignment conducted by Li et al. (2021). The corpus allows preliminary insights into phonetic patterns across a diverse number of languages, with a particularly high representation of low-resource languages, and serves as a foundation from which further investigations can be conducted. The additional processing should also further fulfill the original project's goals of language documentation, preservation, and pedagogical utility for phonetics. It may additionally prove useful for multilingual phone recognition tasks, particularly those involving temporal precision or forced alignment. Our future directions involve standardizing the file names for consistency with the names provided in the UCLA Phonetics Lab Archive recordings and extracting additional data and metadata for increased language coverage and diversity.

### 6. Ethical Considerations and Limitations

The current contribution does not exacerbate any existing ethical concerns with the original public corpus. The UCLA Phonetics Lab Archive was collected several decades ago from around the 1950s to the 1990s by linguistic fieldworkers and associated students. The stated goals included not only linguistic documentation, but also language preservation for the local speech community. The corpus was made available online under a CC-BY-NC 2.0 license from 2006 to 2009 with support from the NSF (Ladefoged and Schuh, 2004). The license restricts the use of the corpus to non-commercial purposes only, preventing potential exploitation through profit.

In using this corpus, we have entrusted the original collectors with having collected speaker consent. The data collection was fieldwork-based in which the fieldworker worked in close communication with the consultant. The corpus occasionally includes speaker demographic information, including year of birth, place of birth, and on the rare occasion, personal names. In some cases, consultants requested that they be personally credited for their contribution (I. Maddieson, personal communication, February 2, 2024). In most cases, however, personally identifying information was not present in the recording or metadata.

In addition, we assume that the original collectors were aware of the language communities' cultural norms regarding recorded language. Some communities may also have concerns regarding digital colonization that could have developed since the time of recording. Requests for data removal from the speakers, their descendants, or their language communities will be honored.

## 7. Acknowledgments

The authors wish to thank Jian Zhu, Claire Moore-Cantwell, Pat Keating, and Ian Maddieson for helpful discussion. This project was supported by SNF Grants PR00P1_208460 to EC and PCEFP1_186841 to SM, and research funding to BP from Ryan Cotterell.